\documentclass[10pt,twocolumn,letterpaper]{article}

\usepackage{cvpr}
\usepackage{times}
\usepackage{amssymb}

\usepackage{float}
\usepackage[utf8]{inputenc}
\usepackage{a4wide}
\usepackage{makeidx}
\usepackage{url}
\usepackage{epsfig}
\usepackage{epstopdf}
\usepackage{doc}
\usepackage{graphicx}
\usepackage{float}
\usepackage{enumerate}
\usepackage{amsmath}
\usepackage{bbm}
\usepackage{multicol, blindtext}
\usepackage{ctable}
\usepackage{textcomp}			

\usepackage[singlelinecheck=false]{caption}

\usepackage[]{footmisc}

\usepackage[breaklinks=true,bookmarks=false]{hyperref}

\cvprfinalcopy 

\def\cvprPaperID{8} 

\begin{document}

\title{Motion Fused Frames: Data Level Fusion Strategy for \\ Hand Gesture Recognition}

\author{Okan K\"op\"ukl\"u \hspace{1cm} Neslihan K\"ose \hspace{1cm} Gerhard Rigoll\\
	Institute for Human-Machine Communication\\
	Technical University of Munich, Germany\\
	{\tt\small \{okan.kopuklu, neslihan.koese, rigoll\}@tum.de}
}

\maketitle
\begin{abstract}
Acquiring spatio-temporal states of an action is the most crucial step for action classification. In this paper, we propose a data level fusion strategy, Motion Fused Frames (MFFs), designed to fuse motion information into static images as better representatives of spatio-temporal states of an action. MFFs can be used as input to any deep learning architecture with very little modification on the network. We evaluate MFFs on hand gesture recognition tasks using three video datasets - Jester, ChaLearn LAP IsoGD and NVIDIA Dynamic Hand Gesture Datasets - which require capturing long-term temporal relations of hand movements. Our approach obtains very competitive performance on Jester and ChaLearn benchmarks with the classification accuracies of 96.28\% and 57.4\%, respectively, while achieving state-of-the-art performance with 84.7\% accuracy on NVIDIA benchmark.  
\end{abstract}

\section{Introduction}

\begin{figure}[t!]
	\includegraphics[width=0.5\textwidth]{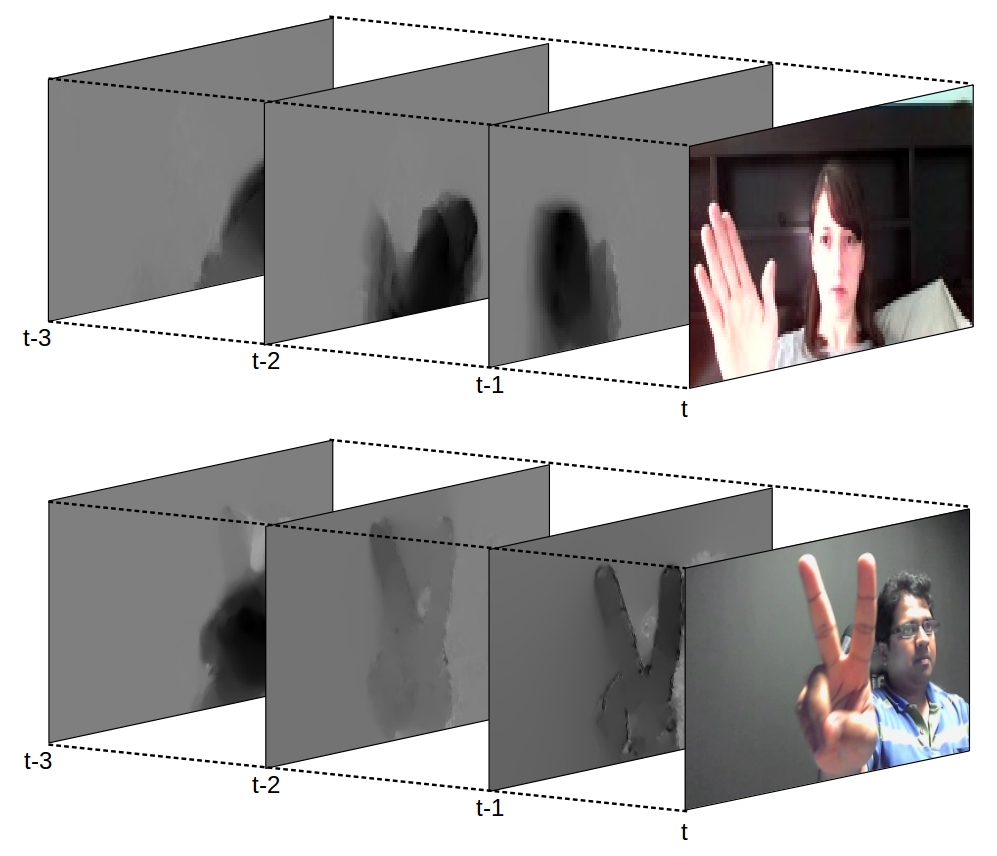}
	\caption{Motion Fused Frames (MFFs): Data level fusion of optical flow and color modalities. Appending optical flow frames to static images makes spatial content aware of which part of the image is in motion and how the motion is performed. Top: 'Swipe-right' gesture. Bottom: 'Showing two fingers' gesture.}
	\label{fig:mff_arch}
\end{figure}

Action and gesture recognition have become very popular topics within computer vision field in the last few years, especially after the application of deep learning in this domain. Similar to other areas of computer vision, the recent work on action and gesture recognition is mainly based on Convolutional Neural Networks (CNNs).

Current CNN architectures are generally designed for static image recognition hence their performances are based on spatial analysis. On the other hand, there is still a gap in the studies applying deep learning for video analysis. Due to the temporal dimension in videos, the amount of data to be processed is high and the models are more complex, which makes the recognition from video data more challenging. Recent studies address this issue and propose approaches analyzing the temporal information in videos \cite{asadi2017survey}.

The temporal information in videos has been analyzed using different modalities such as RGB, depth, infrared and flow images as input. The results show that although each of these modalities provide good recognition performances alone, the fusion analysis of these modalities further increases the recognition performance \cite{Feichtenhofer2016convolutional, simonyan2014two, molchanov2016online}. 

The applied fusion strategy plays a critical role on the performance of multimodal gesture recognition. Different modalities can be fused either on data level, feature level or decision level. Feature and decision level fusions are the most popular fusion strategies that most of the CNNs currently apply \cite{Feichtenhofer2016convolutional, simonyan2014two}. Although they perform pretty well on action and gesture recognition tasks, they have some drawbacks: (i) Usually a separate network must be trained for each modality, which means number of trainable parameters are multiple times of a single network; (ii) at most of the time, pixel-wise correspondences between different modalities cannot be established since fusion is only on the classification scores or on final fully connected layers; (iii) applied fusion scheme might require complex modifications on the network to obtain good results.

The data level fusion is the most cumbersome one since it requires frame registration, which is a difficult task if the multimodal data is captured by different hardwares. However, the drawbacks arising at feature and decision level fusion methods disappear inherently. Firstly, a single network training is sufficient, which reduces the number of parameters multiple times. Secondly, since different modalities are fused at data level, pixel-wise correspondences are automatically established. Lastly, any CNN architecture can be adopted with a very little modification.

In this paper, we propose a data level fusion strategy, Motion Fused Frames (MFFs), using color and optical flow modalities for hand gesture recognition. MFFs are designed to fuse motion information into static images as better representatives of spatio-temporal states of an action. This makes them favorable since hand gestures are composed of sequentially related action states, and slight changes in these states form new hand gestures. To the best of our knowledge, it is the first time that data level fusion is applied for deep learning based action and gesture recognition.

An MFF is generated by appending optical flow frames to a static image as extra channels. The appended optical flow frames are calculated from the consecutive previous frames of the selected static image. Fig. \ref{fig:mff_arch} shows two MFF examples: 'Swipe-right' gesture (top) and 'showing two fingers' gesture (bottom). In the top example of Fig. \ref{fig:mff_arch}, by looking at only static image, one can infer the information of a lady holding her hand upward. However, incorporating optical flow frames into static image brings extra motion information, which shows that the hand is actually moving from left to right making it a swipe-right gesture. 

We evaluated MFFs on three publicly available datasets, which are Jester Dataset \cite{jester}, ChaLearn LAP IsoGD Dataset (ChaLearn) \cite{wan2016chalearn} and NVIDIA Dynamic Hand Gesture Dataset (nvGesture) \footnote{NVIDIA Dynamic Hand Gesture Dataset and ChaLearn LAP IsoGD Dataset are referred as 'nvGesture' and 'ChaLearn' in this paper.}  \cite{molchanov2016online}. Our approach obtains very competitive performance on Jester and ChaLearn datasets with the classification accuracies of 96.28\% ($2^{nd}$ place in the leaderboard) and 57.4\%, respectively, while achieving state-of-the-art performance with 84.7\% accuracy on nvGesture dataset.

The rest of the paper is organized as follows. Section 2 presents the related work in action and gesture recognition that applies deep learning. Section 3 introduces the proposed gesture recognition approach. Section 4 and 5 present the experiments and discussion parts, respectively. Finally, Section 6 concludes the paper.

\section{Related Work}

CNNs have initially been applied for static images, and currently achieve the state-of-the-art performance on object detection and classification tasks \cite{krizhevsky2012imagenet, zhou2014learning, Girshick2014rich}. After they have provided very successful results on static images, they have been extended for recognition tasks on video data \cite{simonyan2014two, Feichtenhofer2016convolutional, Feichtenhofer2016spatiotemporal}.

There are various approaches using CNNs to extract spatio-temporal information from video data. In \cite{Feichtenhofer2016convolutional, simonyan2014two, Karpathy2014largescale, Wang2015towards}, 2D CNNs are used to treat video frames as multi-channel inputs for action classification. 3D CNNs \cite{tran2015learning, tran2017convnet, varol2017long} use 3D convolutions and 3D pooling to capture discriminative features along both spatial and temporal dimensions. Temporal Segment Network (TSN) \cite{wang2016temporal} divides video data into segments and extracts information from color and optical flow modalities for action recognition. Recently, Temporal Relation Network (TRN) \cite{zhou2017temporal} builds on top of TSN to investigate temporal dependencies between video frames at multiple time scales. In \cite{donahue2015long}, the authors propose an architecture, which extracts features from video frames by a CNN and applies LSTM for global temporal modeling. A similar approach \cite{molchanov2016online} proposes a recurrent 3D convolutional neural network for hand gesture recognition, which uses a 3D CNN for the feature extraction part. 

Fusion of information from different modalities is also a common approach in CNNs to increase the recognition performance. There are three main variants for information fusion in deep learning models: data level, feature level and decision level fusions. Within each fusion strategy, still different approaches exists. For instance, for decision level fusion, averaging \cite{simonyan2014two, molchanov2016online}, concatenating \cite{zhou2017temporal} or consensus voting can be applied on the scores of different modalities trained on separate networks. For the feature level fusion case, features from different layers of the CNNs can be fused at different levels \cite{Feichtenhofer2016convolutional}, or different schemes can be proposed as in \cite{miao2017multimodal}, which proposes a canonical correlation analysis based fusion scheme.  

Out of all fusion strategies, data level fusion is the least used one so far since data preparation requires effort especially when different hardwares are used for different modalities. However, it has very critical advantages over feature and decision level fusions like training only a single network, or automatically established pixel-wise correspondence between different modalities. Thus, we applied data level fusion, Motion Fused Frames, to draw the attention to these advantages.    

Considering all the possible CNN architectures, a TSN based approach is selected as building block to leverage MFFs since the architecture proposes to use segmented video clips. Then, MFFs can be used to represent the spatio-temporal state of each segment, which is fused later for classification of the actions. Moreover, MFFs do not necessarily use all the frames in each video segment which is critical for real time applications. 

\begin{figure}[t!]
	\includegraphics[width=0.5\textwidth]{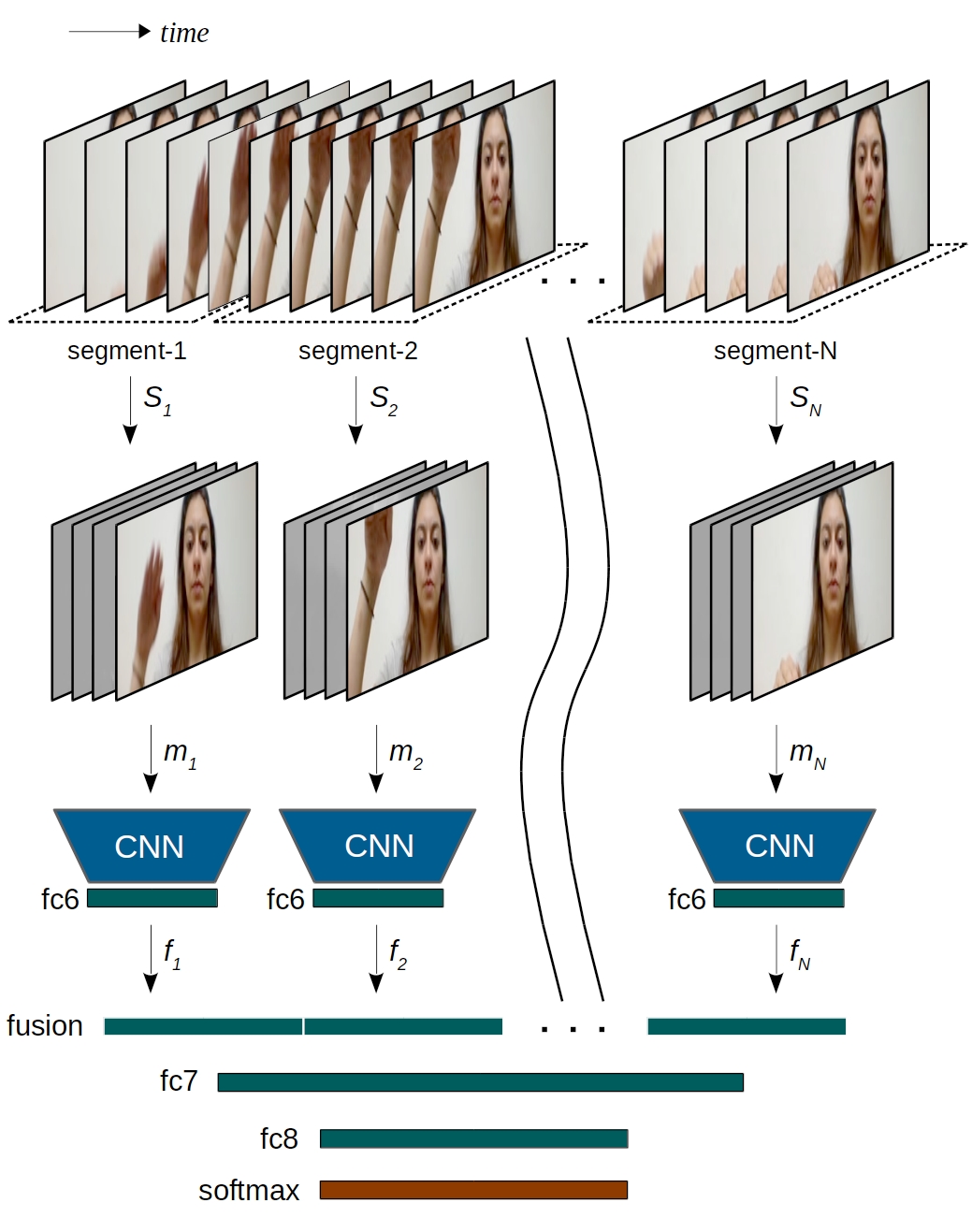}
	\caption{Network Architecture for $N$-segment 3-motion-1-frame MFF ($N$-MFFs-3f1c): One input video is divided into $N$ segments, and equidistant frames are selected from the created segments. 3 optical flow frames calculated from previous frames are appended to RGB frames as extra channels, which forms the Motion Fused Frames (MFFs). Each MFF is fed into a CNN to extract a feature vector representing the spatio-temporal state of the segment. Extracted features are concatenated at the fusion layer and passed to fully connected layers to get class scores.}
	\label{fig:network_arch}
\end{figure}

\section{Approach}

In this section, we describe Motion Fused Frames and the network architecture used for hand gesture recognition. Particularly, we first define MFFs and explain how to form them. Then, we introduce the network architecture which takes advantage of  this data fusion strategy. Finally, we describe the training details on experimented datasets.

\subsection{Motion Fused Frames}

A single RGB image usually contains static appearance information at a specific time instant and lacks the contextual and temporal information about previous and next frames. As a result, single video frames cannot represent the actual state of an action completely. 

Motion fused frames are inspired to be better representatives of action states. As the name implies, they are composed by fusing motion information into static RGB images. Motion information is simply optical flow frames that are calculated from consecutive previous frames, and fusion method is achieved by appending optical flow frames to RGB image as extra channels, as illustrated in Fig. \ref{fig:mff_arch}. Therefore, an MFF contains $(i)$ $spatial$ content contained in $RGB$ channels, and $(ii)$ $motion$ content contained in $optical flow$ channels. Since optical flow images are computed from RGB images, in fact the approach needs only RGB image modality, which avoids the need for enhanced sensors providing several modalities like depth and infrared images.

Blending motion information into contextual information, as in MFFs, ensures pixel-wise correspondence automatically. In other words, spatial content is aware of which part of the image is in motion and how the motion is performed. 

The quality of motion estimation techniques plays a critical role on the performance of gesture recognition. In \cite{varol2017long}, the performance of several optical flow estimation techniques are tested to investigate the dependency of action recognition on the quality of motion estimation. It has been experimentally proved that Brox flow \cite{brox2004high} performs better compared to MPEG flow \cite{kantorov2014efficient} and Farneback \cite{farneback2003two} techniques. Therefore, we have computed horizontal and vertical components of optical flow frames using Brox technique. We have scaled optical flow frames according to the maximum value appeared in absolute values of horizontal and vertical components and mapped discretely into the interval [0, 255]. By means of this step, the range of optical flow frames becomes same as RGB images.

\subsection{Network Architecture}

In this study, we use a deep convolutional neural network architecture applied on segmented video clips, as illustrated in Fig. \ref{fig:network_arch}. The architecture consists of 4 parts: The formation of MFFs, a deep CNN to extract spatio-temporal features from MFFs, fusion of features from different segments and fully connected layers for global temporal modeling, and finally a softmax layer for predicting class-conditional gesture probabilities.

We first divide entire video clip $\mathcal{V}$ into $N$ segments. Each video segment is represented as $S_n \in \mathbb{R}^{w \times h \times c_{rgb} \times m}$ of $m \geq 1$ sequential frames of size $w \times h$ pixels with $c_{rgb}=3$ channels. Then, within segments, equidistant color frames are selected randomly. Each segment is transformed  with $\mathcal{M}$ into an MFF $\textbf{m}_n$ by appending precomputed optical flow frames to the selected color frames:

\begin{equation*}
\begin{split}
& \mathcal{M} : \mathbb{R}^{w \times h \times c_{rgb} \times m} \rightarrow  \mathbb{R}^{w \times h \times c_{mff}}, \\
& where \hspace{0.1cm} m_n = \mathcal{M}(S_n).
\end{split}
\end{equation*}
 
\noindent The number of channels in an MFF can be calculated with $c_{mff} = c_{rgb} + n.c_{flow}$, where $n$ is the number of optical flow frames appended for each segment, and $c_{flow}$ is the number of channels in flow frames that is equal to 2 containing horizontal and vertical components. For instance, an MFF containing 3 flow and 1 color frames, as in Fig. \ref{fig:mff_arch}, has  $c_{mff} = 3 + 3.2 = 9$ channels. Each MFF $\textbf{m}_n$ is then transformed into a feature representation $\textbf{f}_n$ by a CNN $\mathcal{F}$: 
 
\begin{equation*}
\mathcal{F} : \mathbb{R}^{w \times h \times c_{mff}} \rightarrow \mathbb{R}^q, \hspace{0.1cm} where \hspace{0.1cm} f_n = \mathcal{F}(m_n),
\end{equation*}

After extracting feature representations $\textbf{f}_n$ of each MFF, we concatenate them by keeping their order intact: 

\begin{equation*}
(f_1 \oplus f_2 \oplus ... \oplus f_N) \in \mathbb{R}^{N \times q}, 
\end{equation*}

\noindent where $\oplus$ refers to concatenation. We pass this vector into a two-layer multilayer perceptron (MLP). The intuition behind is that the MLP will be able to infer the temporal features from the sequence inherently, without having to know that it is a sequence at all. Finally, a softmax layer is applied to get class-conditional probabilities of each class. ReLU nonlinearity is applied between all convolutional and fully connected layers with the exception of the final fully connected layer which has no nonlinearity.

A network architecture dividing gesture videos into $N$ segments and transforming each segment into an MFF by appending $n$ optical flow frames to 1 color image is referred as $N$-MFFs-$n$f1c.

\subsection{Training Details}

The CNN architecture used to extract features from MFFs is critical for the performance of the overall network, and it has been experimented that deeper architectures like ResNet \cite{he2016deep} performs slightly better results. However, our aim is to evaluate the effectiveness of the proposed data level fusion strategy, Motion Fused Frames, in hand gesture recognition. Therefore, following the design choices of \cite{wang2016temporal}, we adopted Inception with Batch Normalization (BN-Inception) \cite{ioffe2015batch} pretrained on ImageNet as baseline architecture due to its good balance between accuracy and efficiency. We also apply the same training strategies of partial-BN (freezing the parameters of all Batch Normalization layers except the first one) and adding an extra dropout layer after the global pooling layer in BN-Inception architecture. For fc6, fc7 and fc8 layers in Fig. \ref{fig:network_arch}, we used one-layer MLPs with 256, 512 and class-number units, respectively.

For Jester dataset, we modify the weights of first convolution layer of pretrained BN-Inception model to accommodate MFFs. Specifically, the weights across the RGB channels are averaged and replicated through the appended optical flow channels. For ChaLearn and nvGesture datasets, training is started with the pretrained models on Jester dataset. 

\textbf{Learning.} We use stochastic gradient descent (SGD) applied to mini-batch of 32 videos with standard categorical cross-entropy loss. The momentum and weight decay are set to 0.9 and $5 \times 10^{-4}$, respectively. The learning rate is initialized with $1 \times 10^{-3}$ for all the experiments. For Jester dataset, the learning rate is reduced twice with a factor of $10^{-1}$ after $25^{th}$ and $40^{th}$ epochs and optimization is completed after 5 more epochs. For ChaLearn dataset, the learning rate is reduced twice with a factor of $4^{-1}$ after $15^{th}$ and $30^{th}$ epochs and optimization is completed after 10 more epochs. Finally, for nvGesture dataset, the learning rate is reduced twice with a factor of $4^{-1}$ after $40^{th}$ and $80^{th}$ epochs and optimization is completed after 20 more epochs. These training rules are applied for the 8-MFFs-3f1c architecture, and approximately same for the other architectures. 

\textbf{Regularization.} We apply several regularization techniques to reduce over-fitting. Weight decay ($\gamma = 5 \times 10^{-4}$) is applied on all parameters of the network. We use a dropout layer after the global pooling layer (before fc6 in Fig. \ref{fig:network_arch}) of BN-Inception architecture. For Jester dataset, dropout ratio in this layer is kept at 0.8 throughout whole training process. However, over-fitting is much more severe for ChaLearn and nvGesture datasets since average number of training samples per class is much smaller compared to Jester dataset (4391, 144 and 42 training samples per class for Jester, ChaLearn and nvGesture datasets, respectively). Therefore, we apply an additional dropout layer after fc7 layer for these datasets. The dropout ratio is initialized with 0.5 for both dropout layers and increased to 0.8 and 0.9 when the learning rates are reduced. Gradual increase of dropout ratio helps faster convergence while keeping over-fitting in control, which helps to save considerable amount of training time.    

\textbf{Data Augmentation.} Various data augmentations steps are applied in order to increase the diversity of the training videos: $(a)$ Random scaling ($\pm 20\%$), $(b)$ random spatial rotation ($\pm 20$\textdegree), $(c)$ spatial elastic deformation \cite{simard2003best} with pixel displacement of $\alpha$ = 15 and standard deviation of the smoothing Gaussian kernel $\sigma$ = 20 pixels (applied with probability 50\%), $(d)$ random cropping, scale jittering and aspect ratio jittering as in \cite{wang2016temporal}, $(e)$ flipping horizontally with probability 50\% (for only ChaLearn dataset), $(f)$ temporal scaling ($\pm 10\%$) and jittering ($\pm$ 2 frames) (for only nvGesture dataset). All these data augmentation steps are applied online and the input is finally resized to 224 × 224 for network training. 

\textbf{Implementation.} We have implemented our approach in PyTorch \cite{paszke2017automatic} with a single Nvidia Titan Xp GPU. We make our code publicly available \footnote{https://github.com/okankop/MFF-pytorch} for reproducibility of the results.

\section{Experiments}

The performance of the proposed approach is tested on three publicly available datasets: Jester dataset, Chalearn LAP RGB-D Isolated Gesture dataset and NVIDIA Dynamic Hand Gesture dataset. For the evaluation part, center cropping with equidistant frames (middle frame in each segment) in the videos are used for all the datasets.

\subsection{Results Using Jester Dataset}

\begin{table}[t]
	\centering
	\begin{tabular}{lcc}
		\specialrule{.1em}{.5em}{.5em}
		\textbf{Model}                         & \textbf{Top1 Acc.(\%)} & \textbf{Top5 Acc.(\%)} \\ 
		\specialrule{.1em}{.3em}{.3em}
		1-MFFs-0f1c							    & 63.60           & 92.44            \\
		1-MFFs-1f1c                             & 72.83           & 93.96            \\
		1-MFFs-2f1c                             & 73.66           & 94.10            \\
		1-MFFs-3f1c                             & 74.09           & 94.17            \\
		1-MFFs-5f1c                             & 78.39           & 95.84            \\
		1-MFFs-7f1c                             & 81.15           & 96.69            \\
		1-MFFs-9f1c                             & 82.69           & 97.06            \\
		1-MFFs-11f1c                            & 82.93           & 97.07            \\
		\specialrule{.1em}{.3em}{.3em}
		2-MFFs-0f1c							    & 75.65           & 94.40            \\
		2-MFFs-3f1c                             & 84.22           & 97.84            \\
		4-MFFs-3f1c                             & 92.18           & 99.41            \\
		6-MFFs-3f1c                             & 94.72           & 99.66            \\
		8-MFFs-3f1c                             & 95.36           & 99.75            \\
		10-MFFs-3f1c                            & 95.12           & 99.69            \\
		12-MFFs-3f1c                            & 94.73           & 99.69            \\ 
		\specialrule{.1em}{.3em}{.3em}
		8-MFFs-0f1c 						    & 92.90           & 99.41            \\
		8-MFFs-1f1c                             & 94.20           & 99.61            \\
		8-MFFs-2f1c                             & 94.67           & 99.62            \\
		8-MFFs-3f1c                             & 95.36           & 99.75            \\
		\specialrule{.1em}{.3em}{.3em}
		\begin{tabular}[c]{@{}l@{}}\textbf{8-MFFs-3f1c}\\ \hspace{0.3cm}\textbf{(5 crop)}\end{tabular} & \textbf{96.33} & \textbf{99.86}     \\
		\specialrule{.1em}{.3em}{.3em}
	\end{tabular}
	\caption{Results on the validation set of Jester dataset V1}
	\label{tab:jester}
\end{table}

Jester dataset is a recent video dataset for hand gesture recognition. It is a large collection of densely-labeled video clips that shows humans performing pre-defined hand gestures in front of a laptop camera or webcam with a frame rate of 30 fps. There are in total 148,092 gesture videos under 27 classes performed by a large number of crowd workers. The dataset is divided into three subsets: training set (118,562 videos), validation set (14,787 videos), and test set (14,743 videos). 

We initially investigated the effects of the number of appended optical flow frames on the performance of single segment architectures (1-MFFs-$n$f1c). So, we took the complete gesture videos as one segment and tried to classify them using a single RGB image with varying number of appended optical flow frames. We started with 0 optical flow frames and gradually increased it to 11. The results in the first part of Table \ref{tab:jester} show that every extra optical flow frame improves the performance further (from 63.60\% to 82.93\%). The performance boost is significant for the very first optical flow frame with around 9\% accuracy gain.

\begin{table}[t]
	\centering
	\begin{tabular}{lc}
		\specialrule{.1em}{.5em}{.5em}
		\textbf{Model}                           & \textbf{Top1 Acc.(\%)} \\ 
		\specialrule{.1em}{.3em}{.3em}
		C3D							             & 94.62                  \\
		Multiscale TRN \cite{zhou2017temporal}   & 94.78                  \\
		SJ			                             & 94.87                  \\
		Guangming Zhu                            & 95.01                  \\
		DIN 		                             & 95.31                  \\
		NUDT\_PDL 	                             & 95.34                  \\
		MFNet                       			 & 96.22                  \\
		\textbf{8-MFFs-3f1c}                     & \textbf{96.28}         \\
		\textbf{DRX3D}	                         & \textbf{96.60}         \\
		\specialrule{.1em}{.3em}{.3em}
	\end{tabular}
	\caption{Results on the test set of Jester dataset V1}
	\label{tab:jester_test}
\end{table}

\begin{table}[b]
	\centering
	\begin{tabular}{lcc}
		\specialrule{.1em}{.5em}{.5em}
		\textbf{Method}                 & \textbf{Modality} & \textbf{Val. Acc.(\%)} \\ \specialrule{.1em}{.3em}{.3em}
		8-MFFs-0f1c                     & RGB        & 41.3           \\
		8-MFFs-1f1c	                    & RGB + Flow & 48.4           \\
		8-MFFs-2f1c	                    & RGB + Flow & 50.0           \\
		\textbf{8-MFFs-3f1c}	        & RGB + Flow & \textbf{56.9}  \\ 
		\specialrule{0.1em}{.5em}{0em}
	\end{tabular}
	\caption{Results on the validation set of ChaLearn dataset.}
	\label{tab:val_chalearn}
\end{table}

\begin{table}[b]
	\centering
	\begin{tabular}{lcc}
		\specialrule{.1em}{.5em}{.5em}
		\textbf{Method}                 & \textbf{Modality} & \textbf{Test Acc.(\%)} \\ \specialrule{.1em}{.3em}{.3em}
		8-MFFs-0f1c                     & RGB        & 42.8 \\
		8-MFFs-1f1c	                    & RGB + Flow & 53.7 \\
		8-MFFs-2f1c	                    & RGB + Flow & 53.9 \\
		\textbf{8-MFFs-3f1c}	        & RGB + Flow & \textbf{56.7} \\ 
		\specialrule{0.1em}{.5em}{0em}
	\end{tabular}
	\caption{Results on the test set of ChaLearn dataset.}
	\label{tab:test_chalearn}
\end{table}
	
Secondly, we analyzed the effects of number of segment selection for gesture videos. Fixing the number of appended optical frames to 3, we have experimented the 2, 4, 6, 8, 10 and 12-MFFs architectures. The results in the second part of Table \ref{tab:jester} show that the performance increases as we increase the number of segments until reaching to the 8 segmented architecture. Then the performance decreases gradually as we keep incrasing the segment number. In this analysis, it is found that 8 segmented architecture performs best.

Lastly, we analyze the effects of the number of appended optical flow frames on the best performing segment size (8-MFFs-$n$f1c) by varying the number of optical flow frames from 0 to 3. Results in the last part of Table \ref{tab:jester} show that every extra optical flow frame again boosts the performance further. However, the performance boost is more significant for smaller segment architectures like 2-MFFs or 1-MFFs. Out of all models, 8-MFFs-3f1c with 5-crop data augmentation shows the best performance.

We evaluate the 8-MFFs-3f1c architecture on the test set and submit our predictions to the  official leaderboard of the Jester dataset \cite{jester}. At the submission time, our approach is in the second place as shown in Table \ref{tab:jester_test}.


\subsection{Results Using ChaLearn Dataset}

\begin{table}[b]
	\centering
	\begin{tabular}{lcc}
		\specialrule{.1em}{.5em}{.5em}
		\textbf{Method}                             & \textbf{Modality}  & \textbf{Acc.(\%)}  \\ \specialrule{.1em}{.3em}{.3em}
		8-MFFs-0f1c         	     	    		& RGB     & 41.36      \\
		ResC3D \cite{miao2017multimodal}            & RGB     & 45.07      \\
		ResC3D \cite{miao2017multimodal}            & Depth   & 48.44      \\
		ResC3D \cite{miao2017multimodal}            & Flow    & 44.45      \\ \specialrule{.1em}{.3em}{.3em}
		Scene Flow \cite{wang2017scene}	     	    & RGBD    & 36.27      \\
		Wang et al. \cite{wang2016large}            & RGBD    & 39.23      \\
		Pyramidal C3D \cite{zhu2016large}           & RGBD    & 45.02      \\
		2SCVN+3DDSN	\cite{duan2017unified}          & RGBD    & 49.17      \\
		32-frame C3D \cite{7899602}                 & RGBD    & 49.20      \\ 
		C3D+LSTM \cite{zhu2017multimodal}           & RGBD    & 51.02      \\ \specialrule{.1em}{.3em}{.3em}
		8-MFFs-3f1c	  		                        & RGB + Flow & 56.9     \\
		\begin{tabular}[c]{@{}l@{}}\textbf{8-MFFs-3f1c}\\ \hspace{0.3cm}\textbf{(5 crop)}\end{tabular} & RGB + Flow & \textbf{57.4}     \\
		\specialrule{.1em}{.3em}{.3em}
		Zhang et al. \cite{zhang2017learning}	    & RGBD + Flow  & 58.65 \\
		Wang et al. \cite{wang2017large}		    & RGBD + Flow  & 60.81 \\
		\textbf{ResC3D} \cite{miao2017multimodal}   & RGBD + Flow  & \textbf{64.40} \\
		\specialrule{0.1em}{.5em}{0em}
	\end{tabular}
	\caption{Comparison with state-of-the-art on ChaLearn dataset in validation accuracy.}
	\label{tab:benchmark_chalearn}
\end{table}

This database includes 47933 presegmented RGB-D gesture videos each representing one gesture only. There are 249 gesture classes performed by 21 different individuals. The database has been divided into three sub-datasets having 35878, 5784 and 6271 videos for training, validation and testing, respectively. Videos are captured by a Kinect device with a frame rate of 10 fps. 

Experiments on Jester dataset proved that applying MFFs on 8 segmented videos performs better than applying smaller segments. Therefore, we have experimented MFFs on 8 segmented videos with varying the number of optical flow frames. Acquired results for models 8-MFFs-$n$f1c, where $n$ ranges from 0 to 3, are given in Table \ref{tab:val_chalearn} and Table \ref{tab:test_chalearn} for validation and test sets, respectively. Compared to Jester dataset, there is a remarkable performance boost (accuracy gain of 15.6\% and 13.9\% for validation and test sets, respectively) as the number of optical flow frames increases. It must be noted that created MFFs represents a larger time span for ChaLearn dataset since frames are captured with a rate of 10 fps. This gives an intuition that acquired performance at Jester dataset can also be boosted by appending flow frames from earlier timestamps. However, we leave this issue as a future research work.  

Best performing model (8-MFFs-3f1c) is compared with several state-of-the-art methods. According to Table \ref{tab:benchmark_chalearn}, best results are reported in case three modalities are used at the same time. Our approach performs better than most of the approaches reported in the table without using the depth modality, which is a significant advantage of the proposed approach.

\subsection{Results Using nvGesture Dataset}

nvGesture is a dataset of 25 gesture classes, each intended for human-computer interfaces and recorded by multiple sensors and viewpoints. There are 1532 weakly segmented videos in total, which are performed by 20 subjects at an indoor car simulator with both bright and dim artificial lighting. The dataset is randomly split by subjects into training (70\%) and test (30\%) sets, resulting in 1050 training and 482 test videos. Videos are captured by a SoftKinetic DS325 sensor with a frame rate of 30 fps. Since the gesture videos are weakly segmented - some parts of the videos do not contain gesture - we cropped the first and the last 10 frames and used the center 60 frames for the test set evaluation, where the gesture is occurring at most of the time.  

\begin{table}[h]
	\centering
	\begin{tabular}{lcc}
		\specialrule{.1em}{.5em}{.5em}
		\textbf{Method}                             & \textbf{Modality}  & \textbf{Acc. (\%)}  \\ 
		\specialrule{.1em}{.3em}{.3em}
		HOG+HOG$^2$ \cite{ohn2014hand}              & RGB            & 24.5         \\
		\begin{tabular}[c]{@{}l@{}}Spatial stream\\ \hspace{0.3cm}CNN \cite{simonyan2014two}\end{tabular} & RGB   & 54,6 \\
		iDT-HOG \cite{wang2016robust}               & RGB            & 59.1         \\
		C3D	\cite{tran2015learning}                 & RGB            & 69.3         \\
		R3DCNN \cite{molchanov2016online}           & RGB            & 74.1         \\ 
		\specialrule{.1em}{.3em}{.3em}
		iDT-HOF \cite{wang2016robust}               & Flow         & 61.8           \\
		\begin{tabular}[c]{@{}l@{}}Temporal stream\\ \hspace{0.3cm}CNN \cite{simonyan2014two}\end{tabular}   & Flow   & 68,0 \\
		iDT-MBH \cite{wang2016robust}		        & Flow         & 76.8           \\
		R3DCNN \cite{molchanov2016online}		    & Flow         & 77.8           \\ 
		\specialrule{.1em}{.3em}{.3em}
		\begin{tabular}[c]{@{}l@{}}Two stream\\ \hspace{0.3cm}CNN \cite{simonyan2014two}\end{tabular}   & RGB + Flow   & 65,6 \\
		iDT \cite{wang2016robust}                   & RGB + Flow & 73.4             \\
		R3DCNN \cite{molchanov2016online}           & RGB + Flow & 79.3             \\
		6-MFFs-3f1c	  		                        & RGB + Flow & 82.4             \\
		\textbf{8-MFFs-3f1c}	  		            & RGB + Flow & \textbf{84.7}    \\
		\specialrule{.1em}{.3em}{.3em} 
		R3DCNN \cite{molchanov2016online}           & all modalities* & 83.8       \\ 
		\specialrule{.1em}{.3em}{.3em} 
		\textbf{Human} \cite{molchanov2016online}   & RGB             & \textbf{88.4}       \\
		\specialrule{0.1em}{.5em}{0em}
	\end{tabular}
	\caption{Comparison with state-of-the-art on nvGesture dataset. *All modalities refer to RGB, optical flow, depth, infrared and infrared disparity modalities.}
	\label{tab:benchmark_nvgesture}
\end{table}

Although this training set is a lot smaller compared to other datasets, using pretrained models on Jester dataset helps us to remove the over-fitting impact considerably. In Table \ref{tab:benchmark_nvgesture}, we give the comparison of our approach with the state-of-the-art models. Compared to the popular C3D and two stream CNN architectures, our approach can achieve 14.4\% and 19.1\% accuracy gain, respectively. Our approach performs state-of-the-art performance on this benchmark, although we only use color and optical flow modalities. 

The dataset providers also evaluated the human performance by asking six subjects to label each gesture video in the test set for the color modality. Gesture videos are presented to human subjects in random order and only once to be consistent with machine classifiers. Human accuracy is reported as 88.4\% for color modality.

\section{Discussion}

The proposed approach provides a novel fusion strategy for optical flow and color modalities to represent the states of an action. However, this approach is not restricted to only these modalities. Although depth and infrared (IR) modalities require additional hardware, they do not require extra computation like optical flow. This is advantageous for real time applications, and the proposed approach can be applied for such modalities as well, which is the flexibility of our approach.

At the creation of MFFs, optical flow frames calculated from consecutive previous frames are appended to the selected RGB image. The effect of temporally different selection of optical flow frames (i.e. temporally much earlier or later than chosen RGB image) can also be investigated which would increase the performance further.   

A recent work \cite{zhou2017temporal} proposes an approach which can capture temporal relations at multiple time scales. This approach can be used together with our work to acquire better performance. 

After feature extraction, we have used MLPs to get conditional-class scores of the gesture videos. However, different architectures like Recurrent Neural Networks (RNNs) can perform better results. We plan to analyze all these items as future work.

\section{Conclusion}

This paper presents a novel data level fusion strategy, Motion Fused Frames, by fusing motion information (optical flow frames) into RGB images for hand gesture recognition.

We evaluated the proposed MFFs on several recent datasets and acquired competitive results using only optical flow and color modalities. Our results show that, fusion of more motion information improves the performance further at all cases. The performance increase at the first appended optical flow frame is especially significant.

As a future work, we would like to analyze our approach on different modalities at more challenging tasks requiring human understanding in videos. We intend to find better ways to exploit advantages of data level fusion on CNNs for video analysis.

\section*{Acknowledgements}
We gratefully acknowledge the support of NVIDIA Corporation with the donation of the Titan Xp GPU used for this research.

{\small
\bibliographystyle{ieee}
\bibliography{egbib}
}

\end{document}